\documentclass[10pt,logo,copyright]{nvidiatechreport}
\usepackage{graphicx}

\usepackage{xcolor}         %
\usepackage{wrapfig}

\usepackage[numbers, sort]{natbib}
\definecolor{citecolor}{HTML}{0071BC}
\definecolor{linkcolor}{HTML}{ED1C24}

\makeatletter
\@ifpackageloaded{hyperref}{}{
    \usepackage[pagebackref=false, breaklinks=true, colorlinks, citecolor=citecolor, linkcolor=linkcolor, bookmarks=false]{hyperref}
}
\makeatother

\usepackage[utf8]{inputenc} %
\usepackage[T1]{fontenc}    %
\usepackage{hyperref}       %
\usepackage{url}            %
\usepackage{booktabs}       %
\usepackage{amsfonts}       %
\usepackage{nicefrac}       %
\usepackage{microtype}      %
\usepackage{xcolor}         %

\usepackage{epsfig}
\usepackage{graphicx}
\usepackage{amsmath,mathtools}
\usepackage{amssymb}
\usepackage{amsthm}
\usepackage{nicefrac}       %
\usepackage{microtype}      %
\usepackage{amsmath} 
\usepackage{float}
\usepackage{amsfonts}
\usepackage{bm}
\usepackage{enumitem}
\usepackage{mathtools}
\usepackage{multirow}
\usepackage{siunitx}

\usepackage{fontawesome}
\usepackage{pifont}
\usepackage{color}
\usepackage[algo2e,inoutnumbered,linesnumbered,algoruled,vlined]{algorithm2e}
\usepackage{algpseudocode}
\usepackage{tabularx}
\usepackage{url}
\usepackage{bbm}
\usepackage{threeparttable}
\usepackage{tikz}
\usepackage{wrapfig}
\usetikzlibrary{backgrounds}
\usetikzlibrary{decorations.pathreplacing}
\usepackage[outline]{contour}
\usepackage{subcaption}
\usepackage{wrapfig}

\theoremstyle{definition}

\DeclarePairedDelimiterX{\infdivx}[2]{(}{)}{%
  #1\;\delimsize\|\;#2%
}

\definecolor{darkblue}{RGB}{49,130,189}
\definecolor{stanfordgrey}{RGB}{46,45,41}
\definecolor{cardinalred}{RGB}{253,141,60}

\makeatletter
\DeclareRobustCommand\onedot{\futurelet\@let@token\@onedot}
\def\@onedot{\ifx\@let@token.\else.\null\fi\xspace}

\makeatother

\usepackage[capitalize]{cleveref}

\crefname{section}{\S}{\S\S}
\crefname{subsection}{\S}{\S\S}
\crefname{conj}{Conj.}{Conj.}

\Crefname{assumption}{\textbf{H}\hspace{-3pt}}{\textbf{H}\hspace{-3pt}}
\crefname{assumption}{\textbf{H}}{\textbf{H}}

\crefname{algorithm}{\text{Alg.}}{\text{Alg.}}
\crefname{assumption}{\textbf{H}}{\textbf{H}}
\crefname{equation}{\text{Eq}}{\text{Eq}}
\crefname{definition}{\text{Dfn.}}{\text{Dfn.}}
\crefname{lemma}{\text{Lemma}}{\text{Lemma}}
\crefname{dfn}{\text{Dfn.}}{\text{Dfn.}}
\crefname{thm}{\text{Thm.}}{\text{Thm.}}
\crefname{tab}{\text{Tab.}}{\text{Tab.}}
\crefname{fig}{\text{Fig.}}{\text{Fig.}}
\crefname{table}{\text{Tab.}}{\text{Tab.}}
\crefname{figure}{\text{Fig.}}{\text{Fig.}}

\definecolor{mygreen}{RGB}{159, 200, 59}
\definecolor{myred}{RGB}{223, 135, 102}

\newcommand{\PipeName}{\textit{Asset Harvester}\xspace}

\definecolor{policy_green}{rgb}{0.2, 0.59, 0.2}
\definecolor{policy_yellow}{rgb}{0.8, 0.8.0, 0.2}

\usepackage{amsmath,amsfonts,bm,mathtools}

\def\eqref#1{equation~\ref{#1}}

\def\1{\bm{1}}

\DeclareMathAlphabet{\mathsfit}{\encodingdefault}{\sfdefault}{m}{sl}
\SetMathAlphabet{\mathsfit}{bold}{\encodingdefault}{\sfdefault}{bx}{n}

\def\@onedot{\ifx\@let@token.\else.\null\fi\xspace}

\title{\PipeName: Extracting 3D Assets from Autonomous Driving Logs for Simulation}
\author{%
Tianshi Cao*, Jiawei Ren*, 
Yuxuan Zhang, Jaewoo Seo, Jiahui Huang, Shikhar Solanki, 
Haotian Zhang, Mingfei Guo, Haithem Turki,  Muxingzi Li, 
Yue Zhu, Sipeng Zhang, 
Zan Gojcic, Sanja Fidler, Kangxue Yin*\textdagger 
\\
NVIDIA\footnote{*: Core contributors. †: Project lead.}  
}

\begin{abstract}

Closed-loop simulation is a core component of autonomous vehicle (AV) development, enabling scalable testing, training, and safety validation before real-world deployment. Neural scene reconstruction converts driving logs into interactive 3D environments for simulation, but it does not produce complete 3D object assets required for agent manipulation and large-viewpoint novel-view synthesis.
To address this challenge, we present \PipeName, an image-to-3D model and end-to-end pipeline that converts sparse, in-the-wild object observations from real driving logs into complete, simulation-ready assets.
Rather than relying on a single model component, we developed a system-level design for real-world AV data that combines large-scale curation of object-centric training tuples, geometry-aware preprocessing across heterogeneous sensors, and a robust training recipe that couples sparse-view-conditioned multiview generation with 3D Gaussian lifting. Within this system, SparseViewDiT is explicitly designed to address limited-angle views and other real-world data challenges.
Together with hybrid data curation, augmentation, and self-distillation, this system enables scalable conversion of sparse AV object observations into reusable 3D assets.

Code: \href{https://github.com/nvidia/asset-harvester/}{https://github.com/nvidia/asset-harvester/}

Website: \href{https://research.nvidia.com/labs/sil/projects/asset-harvester/}{https://research.nvidia.com/labs/sil/projects/asset-harvester/}

\end{abstract}
\begin{document}

\maketitle

{\hbadness=10000\abscontent\par}


\section{Introduction}
\label{sec:introduction}
\vspace{-0.04in}

Closed-loop simulation is a critical component in developing autonomous vehicles (AVs), enabling scalable training, testing, and safety validation of the policy models in simulation before deploying them in the real world \cite{wang2023drivingfuture,ding2025understanding,russell2025gaia2}. 
NuRec~\cite{nurec_website}, the latest neural reconstruction engine developed by NVIDIA for closed-loop AV simulation, builds 3D reconstructions of real world drives.
A key limitation of per-scene neural reconstruction methods, such as 3DGUT~\cite{wu20253dgut} used in NuRec, is that they can only reconstruct what has been observed.
Unseen regions remain empty or undefined, which is particularly problematic in closed-loop simulation, where changes in ego trajectory and manipulation of traffic participants (e.g., moving or replacing a vehicle) expose previously unobserved regions of 3D objects.

To address the issue, we propose \textbf{\PipeName}, an image-to-3D model and end-to-end pipeline for reconstructing complete, high-quality 3D assets of AV objects, including vehicles, pedestrians, riders, and other common road objects, from real driving videos.
\PipeName targets a data regime that is common in self-driving datasets: vast multi-camera video logs captured by calibrated sensor suites, often accompanied by paired LiDAR data and (sometimes inaccurate) 3D bounding box annotations.
By cropping object instances and selecting one or a few views per object as input, our goal is to convert raw visual evidence into reusable simulation assets at scale.

This setting introduces several challenges. Objects are often observed from limited and biased viewpoints due to ego motion and traffic flow, and they are frequently occluded by other agents and scene elements. In addition, supervisory signals such as 3D cuboid tracks can be noisy or temporally unstable. Camera calibration and synchronization errors further introduce geometric inconsistencies across views, while uncontrolled outdoor illumination and appearance variation complicate robust texture and geometry recovery. For human subjects, non-rigid deformation and clothing motion further violate rigid-body assumptions, making multi-frame aggregation particularly challenging.

\PipeName addresses the above challenges through careful data curation and model designs that tolerate low-quality inputs and imperfect labels.
\PipeName comprises three components: a data ingestion module that extracts suitable image crops from driving logs at scale; \textit{SparseViewDiT}, a Sana-based~\cite{xie2024sana} diffusion model for multiview image generation; and Object TokenGS, which lifts multiview images into 3D Gaussians and is based on the original TokenGS for scene reconstruction~\cite{tokengs2026}.
Our primary contribution lies not only in the individual modules, but also in the design of an end-to-end system for real-world data. Specifically, we introduce a pipeline for large-scale collection and curation of object-centric training tuples from autonomous driving logs, a geometry-aware preprocessing stage that consolidates heterogeneous sensors for multiview generative training, and robust training recipes for SparseViewDiT and TokenGS-based lifting. Within this system, \textit{SparseViewDiT} is designed for sparse, limited-angle observations, and together with hybrid data curation, augmentation, and self-distillation, enables reconstruction of complete 3D assets for vehicles, pedestrians, riders, and other road objects.
We integrate the pipeline with NVIDIA NCore~\cite{ncore_website} for scalable data ingest and with NuRec for asset insertion, harmonization~\cite{zhang2026diffusionharmonizer}, and closed-loop simulation.
Furthermore, we introduce the \textit{NuRec AV Object Benchmark} to support systematic image-to-3D evaluation under realistic viewpoint bias and sensor noise.

\begin{figure*}[t]
\centering
\includegraphics[width=\textwidth]{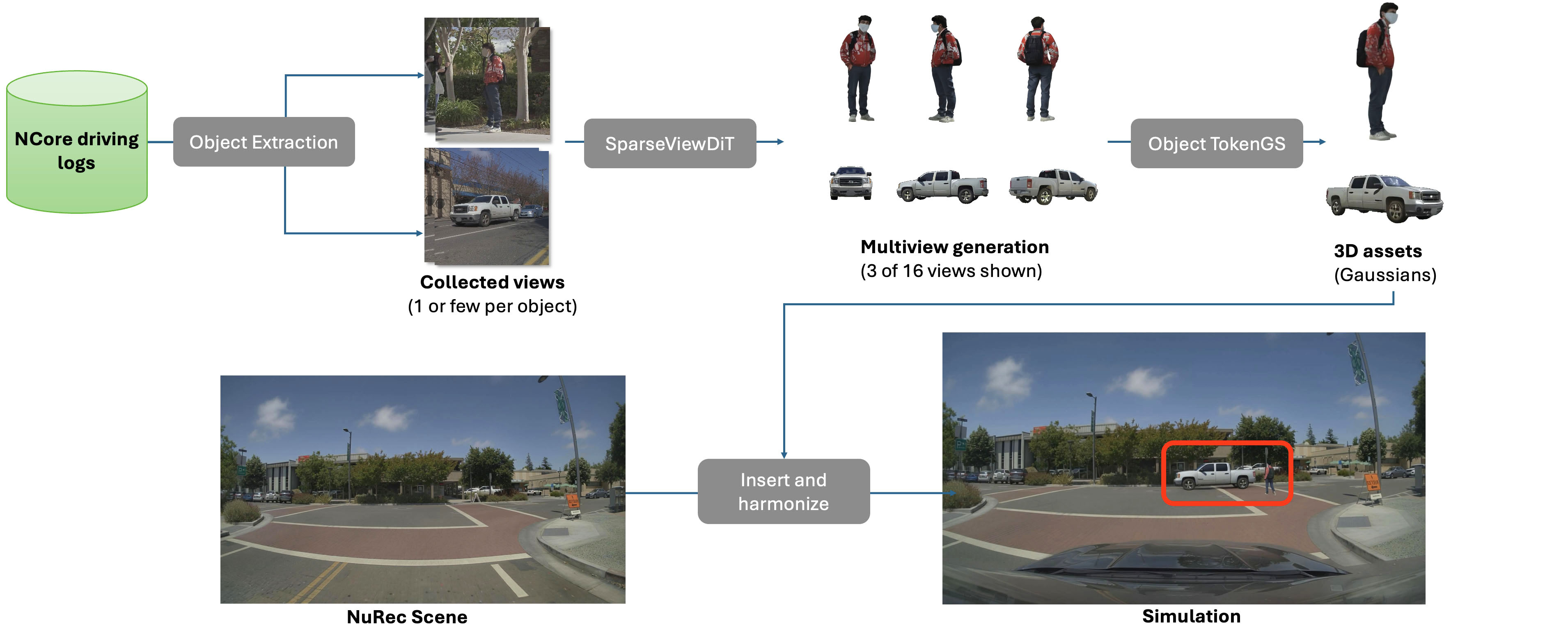}
\caption{Overview of \PipeName. Starting from large-scale AV logs stored in NCore, we crop and rectify object observations, generate multiview images with SparseViewDiT, lift them into 3D assets with an Object TokenGS, and reinsert the assets into scenes with harmonization for closed-loop simulation.}
\label{fig:teaser}
\end{figure*}

\section{{\PipeName}}
\label{sec:model}

Our end-to-end pipeline operates on real autonomous-driving logs in NCore format.
As illustrated in Figure~\ref{fig:teaser}, we extract object-centric views from real-world driving videos and parse their camera parameters.
Given these sparse observations, we follow a standard \emph{multiview generation then 3D lifting} paradigm for sparse-view-conditioned 3D asset generation: SparseViewDiT first synthesizes $16$ uniformly distributed views per object at a given camera FoV and distance, and a feedforward lifting module, referred to as Object TokenGS, then converts the generated set into a compact 3D Gaussian representation with explicit geometry and appearance.
We then perform asset insertion and object replacement followed up by a generative ``harmonizer" that produces photorealistic video frames. Our pipeline enables realistic object-level manipulation, scenario fuzzing and SDG in downstream simulation environments such as NVIDIA NuRec.

\subsection{Data Ingestion}
\label{sec:data_ingestion}

We parse and ingest synchronized multimodal logs containing multi-camera videos, LiDAR sweeps, 3D cuboid tracks, and per-modality timestamps. The implementation details are available in our codebase.

To harvest a complete 3D asset for an object in the recording, we first locate its 3D cuboid track in the ingested logs and align the cuboid track to video frames using timestamp synchronization and camera calibration metadata (intrinsics and extrinsics). We use NVIDIA NCore SDK~\cite{ncore_website} to handle the rolling-shutter effects during projection. Since the original videos are captured by f-theta cameras, we project the 3D cuboid into candidate visible views, crop object-centric patches around the projected cuboid center, and rectify each crop into a canonical pinhole-camera observation~\cite{kannala2006generic}.
To estimate occlusion, we use 3D cuboids of other objects in the same scene: we cast rays from the ego camera to the target cuboid and perform ray--box intersection tests, marking a view as occluded when another cuboid intersects the ray before it reaches the target cuboid. 
To further filter low-quality views, we train a Mask2Former~\cite{cheng2021maskformer} model on object crops from proprietary AV dataset, compare predicted instance masks with projected cuboids, and keep only clean, well-aligned observations. 

Although each object may appear in many consecutive frames, its observed viewpoints along the ego trajectory are often narrow and highly redundant.
Conditioning on all dense frames is inefficient and can hurt learning: near-duplicate inputs contribute little to new geometry, while motion blur, partial occlusion, and calibration/timing noise introduce inconsistent supervision.
To address this, we run farthest-point sampling over camera orientations and keep only a small set of high-quality images with diverse viewing angles as model inputs.
We formulate the problem as a sparse-view-conditioned 3D generation task.
A naive alternative is to first reconstruct a partial object from input videos (e.g., with 3DGS/3DGUT) and then complete it using the partial reconstruction as conditioning, but this is typically suboptimal because partial 3D Gaussian reconstructions tend to overfit visible regions and view-specific artifacts.
We therefore train SparseViewDiT directly with sparse, diverse, high-quality conditioning views, which improves robustness to in-the-wild AV observations and enables reliable geometric completion from limited viewpoints.

\begin{figure*}[t]
\centering
\includegraphics[width=0.7\textwidth]{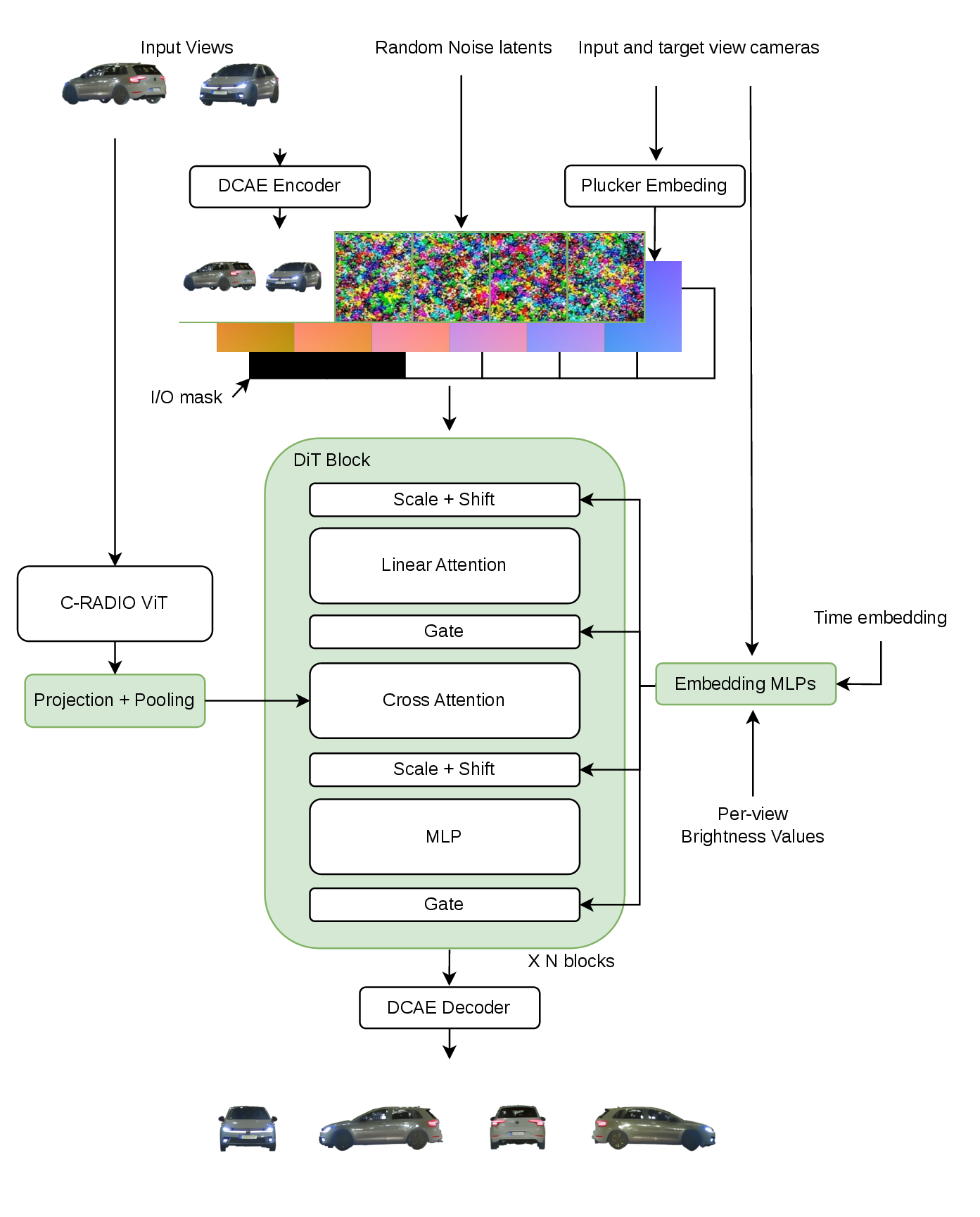}
\caption{Architecture overview of SparseViewDiT for sparse-view-conditioned multi-view generation.}
\label{fig:mvdiffusion}
\end{figure*}

\subsection{SparseViewDiT}
We consider sparse-input novel view synthesis with $v_{in}$ input views and $v_{out}$ target views. Let $X_{in}=\{X_i\}_{i=1}^{v_{in}}$ and $X_{out}=\{X_j\}_{j=1}^{v_{out}}$ denote the corresponding RGB images, and let $\Pi_{in}=\{\pi_i\}_{i=1}^{v_{in}}$ and $\Pi_{out}=\{\pi_j\}_{j=1}^{v_{out}}$ denote their camera parameters. The task is to generate $X_{out}$ conditioned on $(\Pi_{in}, X_{in}, \Pi_{out})$. We operate in a latent-token space: $x_{in}$ and $x_{out}$ denote the (concatenated) latent representations of the input and output views, and we denote the full multiview latent by $x=\mathrm{Concat}(x_{in}, x_{out})$.

We model conditional generation of $x_{out}$ with Flow Matching~\cite{lipman2022flow}. Given conditioning signals $c$=($\Pi_{in}$, $x_{in}$, $\Pi_{out}$), we define a path $p_t(x \mid c)$ from a base noise distribution $p_0(x)=\mathcal{N}(0,I)$ to the conditional data distribution $p_1(x \mid c)$. We use linear (OT) interpolation:

\begin{equation}
    x_t = (1-t)x_0 + t x_1,\quad t\in[0,1],
\end{equation}

where $x_0 \sim p_0$ and $x_1 \sim p_1(\cdot \mid c)$. Generation is obtained by integrating the probability-flow ODE
\begin{equation}
    \frac{dx}{dt} = v_t(x, c),
\end{equation}
with a time-dependent vector field $v_t$. We parameterize $v_t$ by a neural network $v_\theta$ and train it with the conditional flow matching objective (using $v$-prediction):
\begin{equation}
    \mathcal{L}_{FM} = \mathbb{E}_{t, x_0, x_1, c} \left[ || v_\theta(x_t, t, c) - (x_1 - x_0) ||^2 \right]
\end{equation}
where $t$ is sampled uniformly from $[0,1]$. This formulation naturally supports variable numbers of input and target views through the conditioning interface $c$.

Our denoiser $v_\theta$ follows a Diffusion Transformer (DiT) backbone \cite{peebles2023scalable} operating on latent image patches, following the text-to-image design of SANA~\cite{xie2024sana}, with modifications to (i) support sparse input/output configurations and (ii) encode 3D camera geometry explicitly. Our model architecture is provided in Figure~\ref{fig:mvdiffusion}

\paragraph{Flexible Sequence Construction.}
We first encode each masked RGB view into a latent feature map using a VAE~\cite{chen2024deep}, and denote the resulting per-view latents by $x$. We then obtain token sequences by patchifying and linearly embedding these latents, and concatenate tokens from all views along the sequence dimension. With $v_{in}$ input views and $v_{out}$ target views, the total sequence length is $L = (v_{in} + v_{out}) \times L_{patch}$, where $L_{patch}$ is the number of tokens per image, enabling attention layers to directly model cross-view interactions. To distinguish conditioning from generation targets, we concatenate a binary indicator mask $m \in \{0, 1\}^{H \times W}$ channel-wise to the latent feature maps prior to tokenization, where $m=1$ denotes an input view and $m=0$ denotes a target view.

In addition to token-level conditioning, we encode the input images $X_{in}$ with C-Radio~\cite{cradio_website} and provide the resulting image embeddings as context to the DiT cross-attention layers, replacing the text-conditioning interface used in text-to-image diffusion models.

\paragraph{Geometric Conditioning.}
To enforce 3D consistency, we explicitly encode camera geometry using Plücker ray coordinates \cite{mildenhall2020nerf}. For every pixel corresponding to a token, we compute the ray $r = (o, d) \in \mathbb{R}^6$, representing the ray origin and direction. These coordinates are flattened and concatenated to the per-view latent tokens $z$ along the channel dimension. The input to the first transformer layer is thus:
\begin{equation}
    z_{input} = \text{LinearProj}(\text{Concat}(z, r_{plucker}, m))
\end{equation}
Additionally, global camera extrinsics are injected into the model via the time embedding. We map camera parameters to an embedding $e_{cam}$ and inject it into the adaptive layer normalization (adaLN) blocks:
\begin{equation}
    e_{mod} = \text{MLP}(e_t + e_{cam})
\end{equation}
where $e_t$ is the sinusoidal time embedding.

\paragraph{Linear Attention without Positional Embeddings.}
Multiview generation introduces a resolution--memory trade-off: synthesizing $v_{out}$ views at higher spatial resolution increases sequence length and the cost of quadratic attention. We therefore adopt linear attention in the self attention layers and omit absolute positional embeddings, following SANA~\cite{xie2024sana}. Linear attention improves memory efficiency and scalability with sequence length, which facilitates high-resolution multiview synthesis. Instead of relying on token indices to convey spatial layout, the model is informed of per-token geometry via the concatenated Plücker rays and the view-type mask. This design reduces sensitivity to view ordering and accommodates variable numbers of input/output views without reparameterizing positional encodings.


\subsection{3D Lifting}
Given generated multiview outputs $X_{out}=\{X_j\}_{j=1}^{v_{out}}$ and their camera parameters $\Pi_{out}=\{\pi_j\}_{j=1}^{v_{out}}$, 3D lifting reconstructs a 3D asset represented as a set of 3D Gaussians. We denote the lifted representation by
\begin{equation}
    \mathcal{G} = \{ g_k \}_{k=1}^{K},\qquad g_k = (\mu_k, \Sigma_k, \alpha_k, \mathbf{c}_k).
\end{equation}
where $\mu_k\in\mathbb{R}^3$ is the Gaussian center, $\Sigma_k\in\mathbb{R}^{3\times 3}$ is its covariance (parameterized in practice by scale + rotation), $\alpha_k\in[0,1]$ is its opacity, and $\mathbf{c}_k$ denotes view-dependent appearance coefficients (e.g., spherical-harmonic color).

We adopt a feedforward reconstruction network for efficiency and robustness, avoiding per-instance optimization. Following TokenGS~\cite{tokengs2026}, the network uses an encoder--decoder design that predicts a compact set of Gaussian tokens, enabling us to control $K$ independent of image resolution. The model was trained with synthetic supervision and strong camera/image augmentations to improve tolerance to view inconsistency.

\paragraph{Feedforward Reconstruction}
Different from optimization-based 3D reconstruction, feedforward reconstruction learns a mapping
\begin{equation}
    f_\phi:\ (\Pi_{out}, X_{out}) \mapsto \mathcal{G}.
\end{equation}
and is trained such that rendering the predicted Gaussians from each camera matches the corresponding image. Let $\mathcal{R}$ denote differentiable Gaussian splatting rendering. For a camera $\pi_j$, we define the rendered image
\begin{equation}
    \hat{X}_j = \mathcal{R}(\mathcal{G}, \pi_j).
\end{equation}
and optimize a reconstruction loss over output views:
\begin{equation}
    \mathcal{L}_{rec} = \mathbb{E}_{(X_{out},\Pi_{out})}\left[\frac{1}{v_{out}}\sum_{j=1}^{v_{out}} \ell(\hat{X}_j, X_j)\right].
\end{equation}
where $\ell(\cdot,\cdot)$ is a per-pixel loss (e.g., $\ell_1$ / SSIM / perceptual). At inference time, $f_\phi$ produces $\mathcal{G}$ in a single forward pass (seconds per instance).

\paragraph{Model Architecture}
We adopt an encoder--decoder architecture following TokenGS. In encoder-only feedforward reconstruction models, the number of predicted Gaussians grows rapidly with the number of input images and image resolution, leading to substantial redundancy. For example, 16 images at $512	\times 512$ can produce about 4M Gaussians, which is excessive for a single object and degrades rendering efficiency. The encoder--decoder design allows us to control the number of output Gaussians by adjusting the number of query tokens, thereby decoupling the Gaussian count from the number of input pixels. For the encoder, we use a standard ViT-L~\cite{dosovitskiy2020vit}. For the decoder, we use a single cross-attention layer that queries $N$ image features with $M$ Gaussian tokens. Each Gaussian token is decoded into 64 Gaussians. We employ QK-norm and LayerScale in all attention layers.

\paragraph{Augmentation on synthetic data}
Unlike the generative component of our method, reconstruction training requires high-precision, pixel-accurate supervision. Real-world annotations often lack this level of precision due to noise in camera estimation and image corruption. We therefore train the reconstruction model only on synthetic data and apply strong, diverse augmentations to reduce the sim-to-real gap. Specifically, we sample camera poses and intrinsics from real-world data and use them to render the SDG dataset. The field of view ranges from 10$^\circ$ to 40$^\circ$, which is substantially wider than standard fixed-fov rendering. For input images, following LGM~\cite{tang2024lgm}, we apply stronger grid distortion and camera perturbations.


\section{Data Curation and Training}
\label{sec:data}

Learning multi-view image generation requires paired observations across viewpoints, yet such real-world paired data are scarce in autonomous driving. Although multi-view sets can be rendered from synthetic 3D assets, they often exhibit a substantial domain gap relative to real sensor imagery.
To bridge this gap, we curate a hybrid dataset that combines object-centric multi-view observations from real driving videos, synthetic data from in-house domain-specific assets, reconstructions derived from real data, and a broad-coverage Objaverse~\cite{objaverse} subset with commercially viable Creative Commons licenses, all rendered with real world camera and pose distributions.
Building on this data mixture, we adopt a three-stage training pipeline: (i) general-domain pretraining on the Objaverse subset, (ii) in-domain post-training on real and synthetic data, and (iii) supervised fine-tuning on in-domain synthetic data together with the self-distillation set.

\subsection{In-the-wild data pipeline}

We build the real-world dataset from in-house driving logs stored in NCore format, which provides synchronized multi-sensor recordings (RGB videos and Lidar), camera intrinsics/extrinsics, and 3D detection/tracking annotations for on-road objects.

Following the ingestion procedure in Section~\ref{sec:data_ingestion}, we focus here on collection and curation. From candidate object-centric crops produced by running the ingestion pipeline over a large set of AV driving sessions, we group observations of the same tracked object across cameras and timestamps to form per-instance multi-view pools.
We then apply quality and coverage filters. We discard candidates with low resolution, severe boundary truncation, heavy overlap with other 3D boxes, or camera-to-object distances outside a valid range. For each instance, we run farthest-point sampling on camera orientations and retain at most 32 images, maximizing angular diversity while reducing near-duplicate views.

Finally, we apply Qwen2.5-VL~\cite{qwen2.5-VL} to remove blurry or heavily occluded samples. We run background segmentation on the remaining images using the Mask2Former model described in Section~\ref{sec:data_ingestion}, followed by VLM-based checks and manual review. The final curated collection contains 278k multi-image sets of dynamic street-scene objects, including automobiles, pedestrians, riders, and construction machinery.

\subsection{3D synthetic data}
We render synthetic 3D assets from various sources to supply relevant 3D consistent multi-view image sets to train \PipeName.

\paragraph{Objaverse commercially viable subset}
For general domain 3D supervision, we use 80k assets from an Objaverse subset with commercially viable licenses. For each asset, we render two camera sets. The first set is randomly sampled from the camera distribution estimated from the real driving data. The second set contains 16 views whose camera FoVs and distances are sampled from the same real data distributions, with azimuths fixed at uniformly spaced angles around the object and elevations fixed at 0°. This design provides both realistic viewpoint statistics and consistent canonical coverage for stable multi-view 3D lifting training.

\paragraph{In-domain vehicle data generation}
To provide high-quality in-domain vehicle observations with consistent geometry, we render a synthetic vehicle dataset using in-house 3D assets. The asset library contains approximately 200 vehicles represented in USD format with detailed part-level components. During rendering, we randomize vehicle appearance and environmental conditions using Omniverse Replicator~\cite{omniverse_replicator}, including body color, material properties, and scene lighting.
For each vehicle asset, we generate 20 samples with randomized camera viewpoints and fields of view, following pose distributions estimated from real driving data. This procedure yields diverse multi-view observations while preserving physically consistent geometry and realistic sensor perspectives. The resulting dataset provides strong supervision for vehicle categories commonly encountered in autonomous-driving environments.

\paragraph{In-domain human data generation.}
We generate a synthetic human dataset using RenderPeople assets~\cite{renderpeople} and public-domain HDRI backgrounds from PolyHaven~\cite{polyhaven}. We also use an in-house motion dataset (walk/idle/dancing, etc.) and camera pose distributions estimated from real driving videos and logs. Each sample is built from one randomly selected person and consists of two sets of renders: (1) 16 images rendered with predefined camera viewpoints in a T-pose and a random camera FoV, and (2) 16 images rendered with a random animation pose and random camera viewpoints/FoV. Thus, each person--HDR background pair yields 32 images in total. We render 12,800 such samples using NVIDIA Omniverse. 

\paragraph{Self-distillation Data}
After post-training, \PipeName can generate multiview images that largely preserve the detail and quality of a clean input image. However, images from real on-vehicle sensors often suffer from strong motion blur and distortion, which lead to low-quality predictions. To mitigate this issue, we construct a self-distillation dataset by first creating paired low- and high-quality single-view images from our real-world image set, where the high-quality images are enhanced with Qwen-Image-Edit~\cite{wu2025qwenimagetechnicalreport}. We then use \PipeName to generate multiview images conditioned on the high-quality image and form new training pairs consisting of the low-quality single-view image and the corresponding high-quality generated multiview images. After screening the results, we keep up to 1000 samples per object class for training.

\subsection{Training}

\paragraph{Multi-stage Training}
Starting from the pretrained Sana-1.6B model, we employ a multi-stage optimization pipeline designed to progressively align the model with 3D consistent multi-view image generation with realistic appearances for autonomous-driving scenarios. In the first stage, we perform 3D pre-training on Objaverse-rendered data to establish strong general-domain 3D priors and adapt the base model to structured multi-view synthesis. In the second stage, we conduct in-domain post-training on a mixed corpus of real image sets and synthetic 3D-rendered data, which improves robustness to real-world appearance variations while preserving geometric consistency. Finally, after constructing a self-distillation dataset from the post-trained model, we fine-tune on a mixture of self-distillation data and synthetic 3D-rendered data to further enhance view consistency, visual fidelity, and stability under challenging real-sensor artifacts.

\paragraph{Augmentation Strategies}
During post-training and fine-tuning, we apply several augmentations to improve robustness to noisy real-world AV inputs. We synthesize occlusions using existing object masks so the model is regularly exposed to partially visible targets, improving stability under realistic traffic occlusion patterns. We also randomly perturb camera fov and camera extrinsics to reduce sensitivity to noise in geometric conditioning signals. In addition, we apply brightness augmentation before C-RADIO encoding of the image prompt to simulate ISP inconsistency across multi-camera sensor rigs. Together, these augmentations improve robustness to imperfect geometry, appearance shifts, and sensor-domain mismatch.
\section{NuRec AV Object Benchmark}
\label{sec:benchmark}

\subsection{Data and Statistics}

The NuRec AV Object Benchmark contain two complementary parts that reflect both controlled and challenging in-the-wild evaluation settings.

\paragraph{Part A (held-out-view evaluation).} For each object instance, we reserve unseen reference views that are not used as model input. These held-out views provide pseudo-ground-truth targets for quantitative image-to-3D evaluation. In this split, we report pixel correspondence based metrics including the metrics we built on DINOv3 features. The metric design explicitly supports non-rigid humans: when the object is a person, keypoints are detected and a body-part-aware distance is computed.

\paragraph{Part B (no-ground-truth hard split).} This part is intentionally more challenging, with stronger motion blur, heavier occlusion, and narrower view coverage. No reserved reference views are available, so direct ground-truth reconstruction metrics cannot be computed. Instead, we use GPT5.2~\cite{openai2026gpt52} to evaluate how consistent the rendered outputs are with the available input views.

\paragraph{Object taxonomy.}
Both Part A and Part B include the same five object classes: VRU pedestrians, VRU riders (bicycles, motorcycles, and scooters), commercial vehicles (e.g., buses and trucks), consumer vehicles (e.g., sedans, SUVs, and pickups), and other objects (e.g., construction machines, trailers, trash bins, golf carts, etc.). Table~\ref{tab:nurec-sample-counts} summarizes the number of benchmark instances per class and split.

\begin{table}[htbp]
\centering
\caption{Number of samples per class in NuRec AV Object Benchmark.}
\label{tab:nurec-sample-counts}
\begin{tabular*}{\linewidth}{@{\extracolsep{\fill}}lrrr}
\toprule
Class & Part A & Part B & Total \\
\midrule
Consumer vehicles   & 1,472 & 602 & 2,074 \\
Commercial vehicles & 308 & 405 & 713 \\
VRU (pedestrians)     & 330 & 383 & 713 \\
VRU (riders)          & 41 & 30 & 71 \\
Other objects       & 55 & 90 & 145 \\
\midrule
\textbf{Total}      & 2,206 & 1,510 & 3,716 \\
\bottomrule
\end{tabular*}
\end{table}

\subsection{Metrics}

We evaluate generated assets by rendering them from the cameras of preserved views and comparing the renderings against the corresponding reference images. Although we also consider standard metrics such as PSNR, SSIM and LPIPS,  these pixel/perceptual metrics are sensitive to small spatial misalignments (e.g., slight differences in pose, scale, or position), and real-world AV data often contains cuboid noise and imperfect camera calibration that further destabilize pixel-aligned evaluation. We therefore introduce the following high-level metrics to better measure semantic and structural similarity between rendered and preserved images.

\paragraph{Rigid Embedding Distance (ED-R)}

To capture higher-level semantic and structural similarity, we introduce two complementary metrics based on DINOv3~\cite{Simeoni2025DINOv3} embeddings.

For ED-R, we first align the rendered object to the preserved ground-truth (GT) view using foreground masks. Let $B_r$ and $B_{gt}$ be the rendered and GT binary object masks, with centers $\mathbf{c}_r, \mathbf{c}_{gt}$ and mask areas $A_r, A_{gt}$. We translate the rendered view to align mask centers and scale it by
$\sqrt{\frac{A_{gt}}{A_r}}$
so that the rendered mask area matches the GT mask area. Applying this transform to the rendered image and mask gives $\tilde{I}_r$ and $\tilde{B}_r$.

We then extract DINOv3 patch features $\mathbf{F}_r, \mathbf{F}_{gt} \in \mathbb{R}^{C \times H_p \times W_p}$ from $\tilde{I}_r$ and $I_{gt}$. Using the corresponding masks, we query foreground patch features and average them:
\begin{equation}
  \mathbf{e}_r = \frac{1}{|\mathcal{S}_r|}\sum_{(i,j)\in\mathcal{S}_r}\mathbf{F}_{r,:,i,j},
  \quad
  \mathbf{e}_{gt} = \frac{1}{|\mathcal{S}_{gt}|}\sum_{(i,j)\in\mathcal{S}_{gt}}\mathbf{F}_{gt,:,i,j},
\end{equation}
where $\mathcal{S}_r$ and $\mathcal{S}_{gt}$ are foreground patch sets defined by $\tilde{B}_r$ and $B_{gt}$. ED-R is the cosine distance between the pooled embeddings:
\begin{equation}
  d_{\text{ED-R}} = 1 - \frac{\mathbf{e}_r \cdot \mathbf{e}_{gt}}{\|\mathbf{e}_r\| \, \|\mathbf{e}_{gt}\|}.
\end{equation}

\paragraph{Part-aware Embedding Distance (ED-P).}
ED-P is a part-aware version of ED-R and is computed only for pedestrian instances. We run SAM 3D Body~\cite{yang2026sam3dbody} to detect human keypoints and use them to partition each human mask into body parts (head, torso, left/right arms, and left/right legs). DINOv3 patch features are then average-pooled within each part to obtain part embeddings:
\begin{equation}
  \mathbf{e}^k = \frac{1}{|\mathcal{S}_k|}\sum_{(i,j)\in\mathcal{S}_k}\mathbf{F}_{:,i,j},
\end{equation}
where $\mathcal{S}_k$ is the patch set for part $k$.
For each part, we compute cosine distance between rendered and GT embeddings, and define ED-P as the average part distance:
\begin{equation}
  d_{\text{ED-P}} = \frac{1}{|\mathcal{K}|} \sum_{k \in \mathcal{K}} \left(1 - \frac{\mathbf{e}^k_{r} \cdot \mathbf{e}^k_{gt}}{\|\mathbf{e}^k_{r}\| \, \|\mathbf{e}^k_{gt}\|}\right),
\end{equation}
where $\mathcal{K}$ is the set of body parts visible in both rendered and GT masks.

\paragraph{GPT5.2 Pairwise Preference Rate }
For benchmark Part B, where no held-out ground-truth views are available, we perform pairwise comparison between \PipeName and each baseline using GPT5.2. For each comparison, we provide exactly three images: a reference input view (A), one randomly rendered view from one method (B), and one randomly rendered view from the other method (C). We randomize whether \PipeName is assigned to B or C to avoid position bias. 
The GPT5.2 system prompt is shown below:

\begin{center}
\setlength{\fboxsep}{6pt}
\fbox{%
\begin{minipage}{0.95\linewidth}
\fontsize{8.5}{9.5}\selectfont\ttfamily
You will receive exactly three images in this order:\\
- A: reference image (object at the center of the image; may be partially occluded; background was masked out)\\
- B: rendered image of a 3D reconstruction in random viewpoint\\
- C: rendered image of another 3D reconstruction in random viewpoint\\
\\
Which of B or C is overall closer to the object in A (shape and appearance of the object itself)?\\
Ignore differences that are mainly due to random viewpoint, scale, translation, or occlusion of the object.\\
\\
Reply with exactly one bracket containing a single token:\\
- B if B is closer to the reference object\\
- C if C is closer\\
- ERROR if you did not receive exactly three images\\
\\
Format: [X] where X is B, C, or ERROR. No other text.
\end{minipage}%
}
\end{center}

After mapping B/C back to method identities, we report the percentage of valid pairwise comparisons in which \PipeName is preferred over the baseline.

\section{Experiments and Results}

In this section, we first present quantitative results on the benchmark introduced above, including comparisons with image-to-3D baselines, a GPT-based study on the harder split, an ablation on the number of input views, and inference-time measurements. 
We then present qualitative results demonstrating in-the-wild performance, out-of-distribution image generalization, pedestrian animation with our assets, and asset insertion and simulation in NuRec.

\subsection{Quantitative Evaluation and Analysis}

We compare \PipeName against representative image-to-3D baselines, including SAM3D~\cite{sam3dteam2025sam3d3dfyimages}, TRELLIS~\cite{xiang2024trellis}, and Hunyuan3D~\cite{zhao2025hunyuan3d2,hunyuan3d2025hunyuan3d}, on the NuRec AV Object Benchmark.

\paragraph{Comparison Setup.}
For a fair comparison, we use a single RGB input view for all methods. For Asset Harvester (AH), we report two single-view settings: (1) without camera metadata, where we estimate camera parameters using a linear-probing MLP on top of C-Radio features (trained on our training split), and (2) with camera parameters parsed from the AV scene.

\begin{figure*}[t]
\centering
\includegraphics[width=1.0\textwidth]{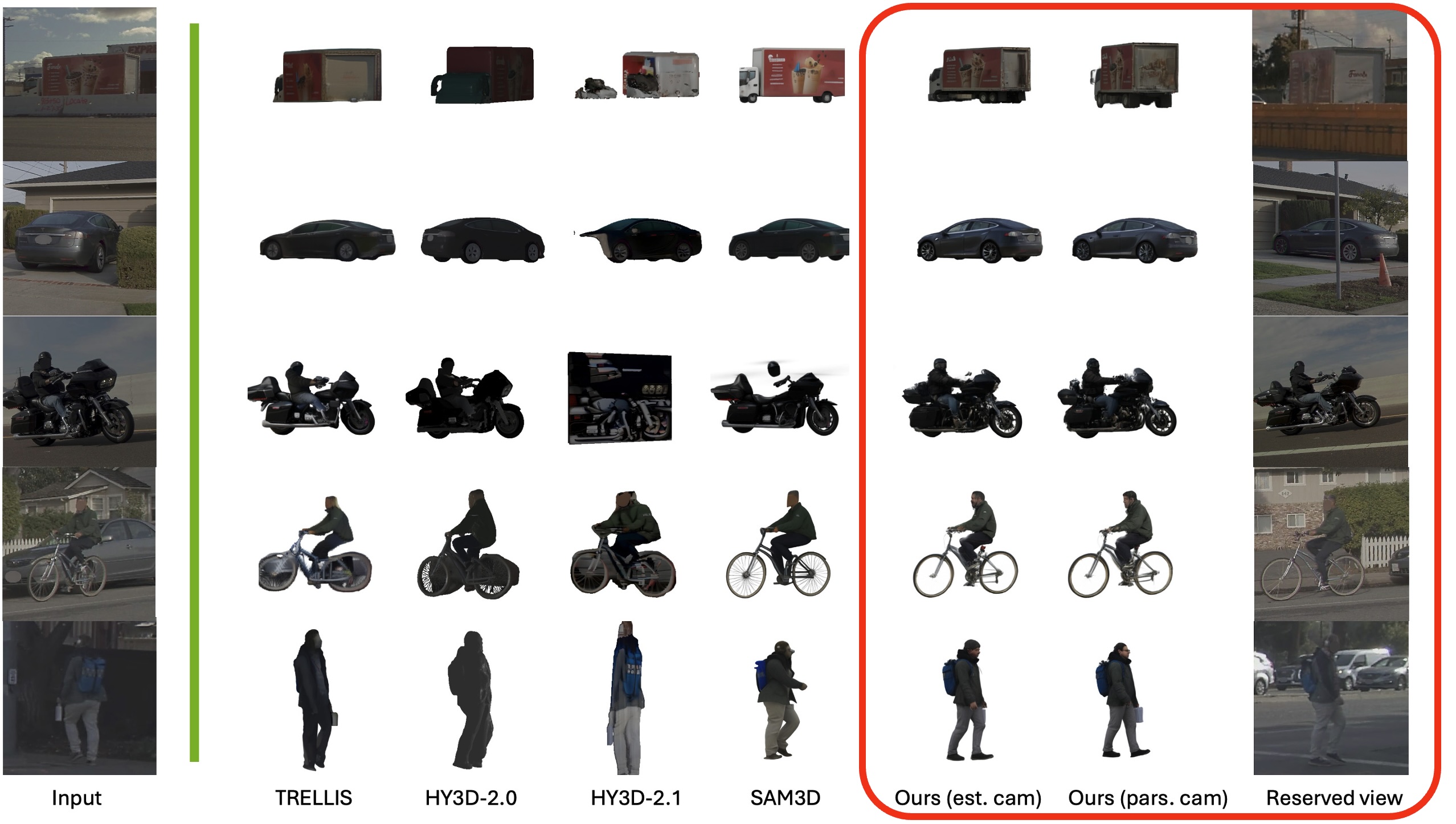}
\caption{Qualitative comparison against image-to-3D baselines (zoom in for details). Asset Harvester is evaluated under two camera settings: estimated camera pose and parsed camera pose.}
\vspace{-0.5em}
\label{fig:qualitative-comparison}
\end{figure*}

\paragraph{Main Results.}
As shown in Table~\ref{tab:main-comparison}, \PipeName achieves the best overall performance across all metrics against reserved views in Part A of the benchmark. The detailed evaluation results for each class are provided in the Appendix. We provide qualitative comparison with a few examples in Figure~\ref{fig:qualitative-comparison}.

\begin{figure*}[p]
\centering
\includegraphics[width=0.9\textwidth,keepaspectratio]{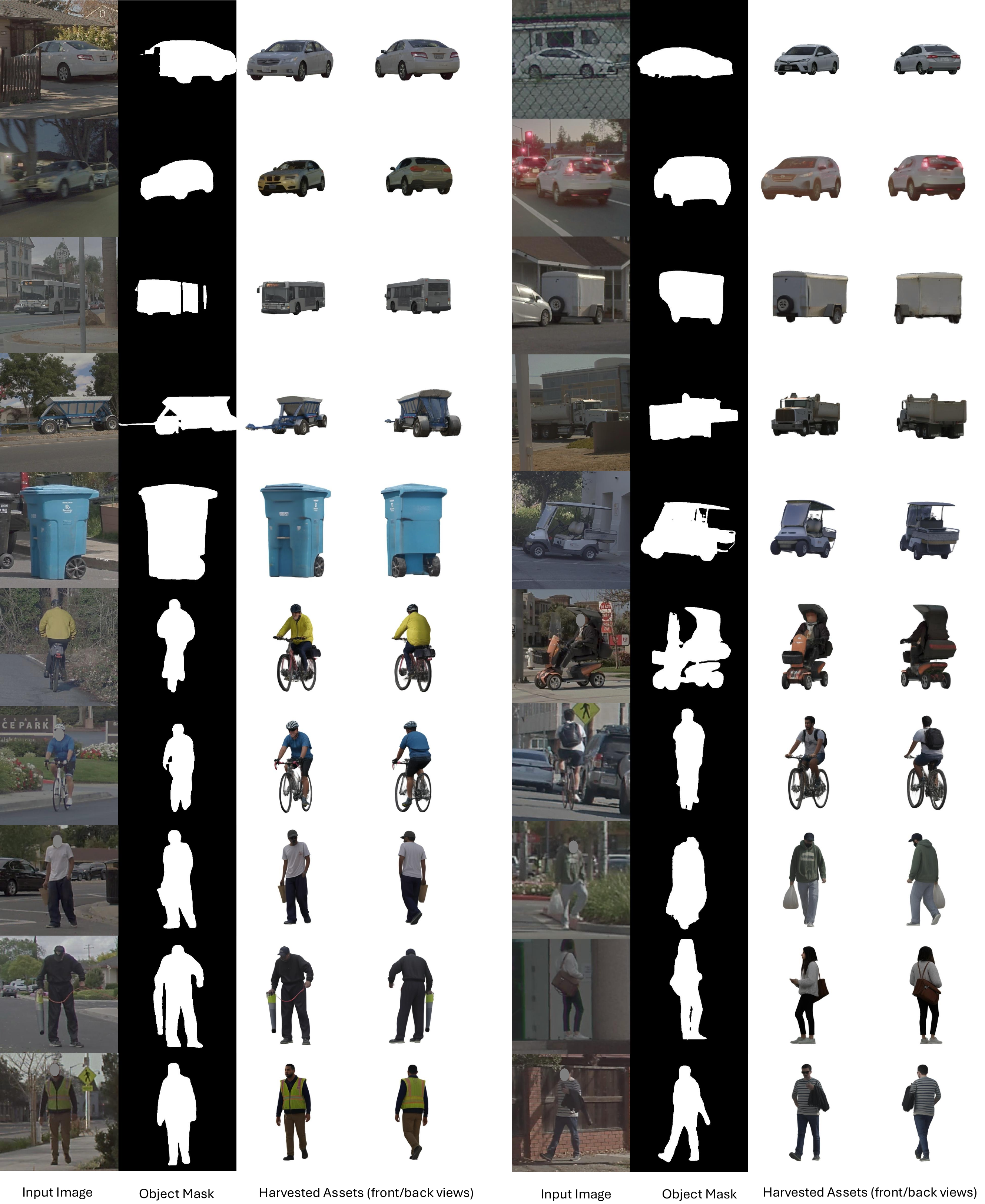}
\caption{In-the-wild qualitative results across diverse object classes, including sedan, bus, trailer, trash bin, truck, rider, and pedestrian. The figure is organized into two result columns per row. In each result column, from left to right, we show the \textbf{single-view} input image, the extracted object mask, and front/back views of the generated 3D object.}
\label{fig:in-the-wild}
\end{figure*}

\begin{table*}[htbp]
\centering
\caption{Quantitative comparison with image-to-3D baselines on Part A of the NuRec AV Object Benchmark. The best and second-best results for each metric are highlighted in bold. ED-P is a non-rigid metric and is computed only for pedestrian instances.}
\label{tab:main-comparison}
\small
\begin{tabular*}{\linewidth}{@{\extracolsep{\fill}}lccccccc}
\toprule
Method & PSNR $\uparrow$ & SSIM $\uparrow$ & LPIPS $\downarrow$ & ED-R $\downarrow$ & ED-P((Ped.-only)) $\downarrow$  \\
\midrule
TRELLIS~\cite{xiang2024trellis} & 20.71 & 0.854 & 0.190  & 0.155 & 0.290  \\
Hunyuan3D-2~\cite{zhao2025hunyuan3d2} & 19.47 & 0.848 & 0.202  & 0.204 & 0.356  \\
Hunyuan3D-2.1~\cite{hunyuan3d2025hunyuan3d} & 18.86 & 0.825 & 0.213 & 0.229 & 0.395  \\
SAM3D~\cite{sam3dteam2025sam3d3dfyimages} & 20.64 & 0.855 & 0.174  & 0.127 & 0.257  \\
\midrule
\textbf{\PipeName} (1V, est. cam pose) & \textbf{21.99} & \textbf{0.861} & \textbf{0.163}  & \textbf{0.104} & \textbf{0.241}  \\
\textbf{\PipeName} (1V, parsed cam pose) & \textbf{22.23} & \textbf{0.864} & \textbf{0.153}  & \textbf{0.098} & \textbf{0.231}  \\
\bottomrule
\end{tabular*}
\end{table*}

\begin{figure*}[t]
\centering
\includegraphics[width=1.0\textwidth,keepaspectratio]{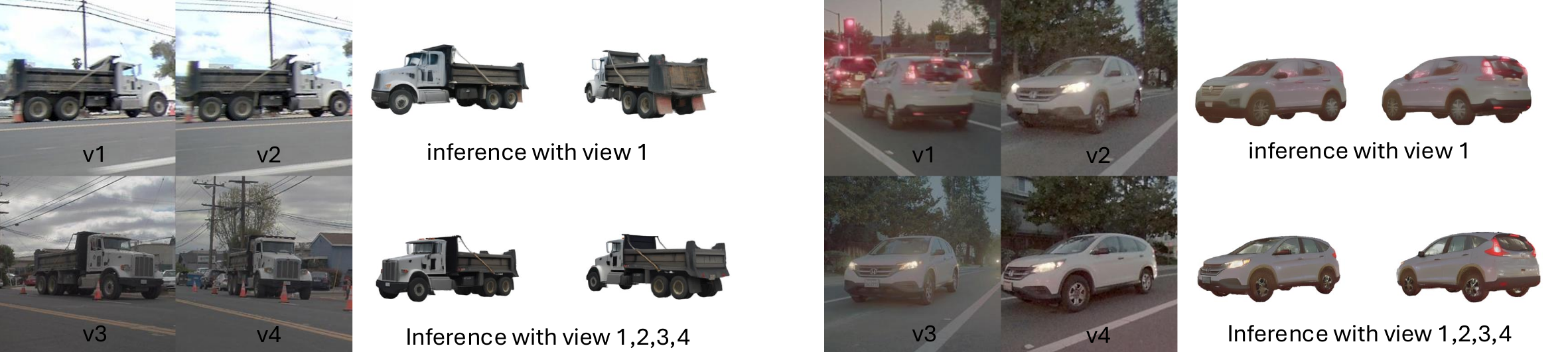}
\caption{Input-view ablation for \PipeName. We compare 1-view input results against 4-view input results. While single-view input already yields plausible reconstructions, additional input views with better object coverage improve sharpness and details.}
\vspace{-1.0em}
\label{fig:view-ablation}
\end{figure*}


\paragraph{GPT-study}
On benchmark Part B, the reserved views are not available due to the limited viewing angles of objects in AV scenes. Therefore, we instead conduct a GPT-study where the evaluator is GPT-5.2, which is the GPT-5.2-based pairwise metric described in Section~\ref{sec:benchmark}. The results are provided in Table~\ref{tab:gpt-study-partb}. Detailed per-class comparisons are provided in the Appendix.

\begin{table*}[htbp]
\centering
\caption{GPT-5.2 pairwise preference rates on Part B of the NuRec AV Object Benchmark. Higher percentages indicate that the corresponding method is preferred more often. AH denotes \PipeName.}
\label{tab:gpt-study-partb}
\small
\resizebox{\linewidth}{!}{%
\begin{tabular}{lcccccccc}
\hline
\textbf{Comparison} & \multicolumn{2}{c|}{\textbf{AH vs Trellis}} & \multicolumn{2}{c|}{\textbf{AH vs HY2}} & \multicolumn{2}{c|}{\textbf{AH vs HY2.1}} & \multicolumn{2}{c}{\textbf{AH vs SAM3D}} \\
\hline
 & AH (\%) & Baseline (\%) & AH (\%) & Baseline (\%) & AH (\%) & Baseline (\%) & AH (\%) & Baseline (\%) \\
1V, est. cam pose &  \textbf{72.9} & 27.1 & \textbf{78.8} & 21.2 & \textbf{76.8} & 23.2 & \textbf{57.3} & 42.7 \\
1V, parsed cam pose &  \textbf{73.9} & 26.1 & \textbf{78.4} & 21.6 & \textbf{75.6} & 24.4 & \textbf{59.8} & 40.2 \\
\hline
\end{tabular}
}
\end{table*}

\paragraph{Inference speed}
We measure the end-to-end inference time of \PipeName on single NVIDIA A100 and H100 GPU and report the results in Table~\ref{tab:inference-time}. 

\paragraph{Ablation: Input Views}
Object views in our benchmark are sampled with a minimum inter-view angle of $15^\circ$, and AV viewpoints are inherently limited. As a result, only a small number of valid frames remain per sample, often just a single view. For the input-view ablation, we therefore select a subset of Part A that contains at least three valid input views.
We report the resulting quality metrics under different input-view settings in Table~\ref{tab:ablation-input-views}.

Figure~\ref{fig:view-ablation} compares reconstructions from 1-view and 4-view inputs. The model already produces reasonable assets from a single image, but when multiple input views provide better object coverage, the generated geometry and appearance are visibly sharper and more detailed.

\begin{table*}[htbp]
\centering
\begin{minipage}[t]{0.45\linewidth}
\centering
\caption{Asset Harvester inference-time }
\label{tab:inference-time}
\small
\begin{tabular*}{\linewidth}{@{\extracolsep{\fill}}lcccr}
\toprule
GPU & Encoding & Diffusion & Lifting & Total \\
\midrule
A100 & 0.51 s & 7.38 s & 3.06 s & 10.95 s \\
H100 & 0.27 s & 3.82 s & 1.69 s & 5.78 s \\
\bottomrule
\end{tabular*}
\end{minipage}\hfill
\begin{minipage}[t]{0.53\linewidth}
\centering
\caption{Ablation on number of input views}
\label{tab:ablation-input-views}
\small
\begin{tabular*}{\linewidth}{@{\extracolsep{\fill}}lccccc}
\toprule
Input & PSNR $\uparrow$ & SSIM $\uparrow$ & LPIPS $\downarrow$ & ED-R $\downarrow$ & \shortstack{ED-P\\(Ped.-only)} $\downarrow$ \\
\midrule
1-view & 21.68 & \textbf{0.855} & 0.156 & 0.098 & 0.219 \\
3-view & \textbf{21.71} & \textbf{0.855} & \textbf{0.152} & \textbf{0.087} & \textbf{0.207} \\
\bottomrule
\end{tabular*}
\end{minipage}
\end{table*}


\begin{figure*}[t]
\centering
\includegraphics[width=0.9\textwidth,keepaspectratio]{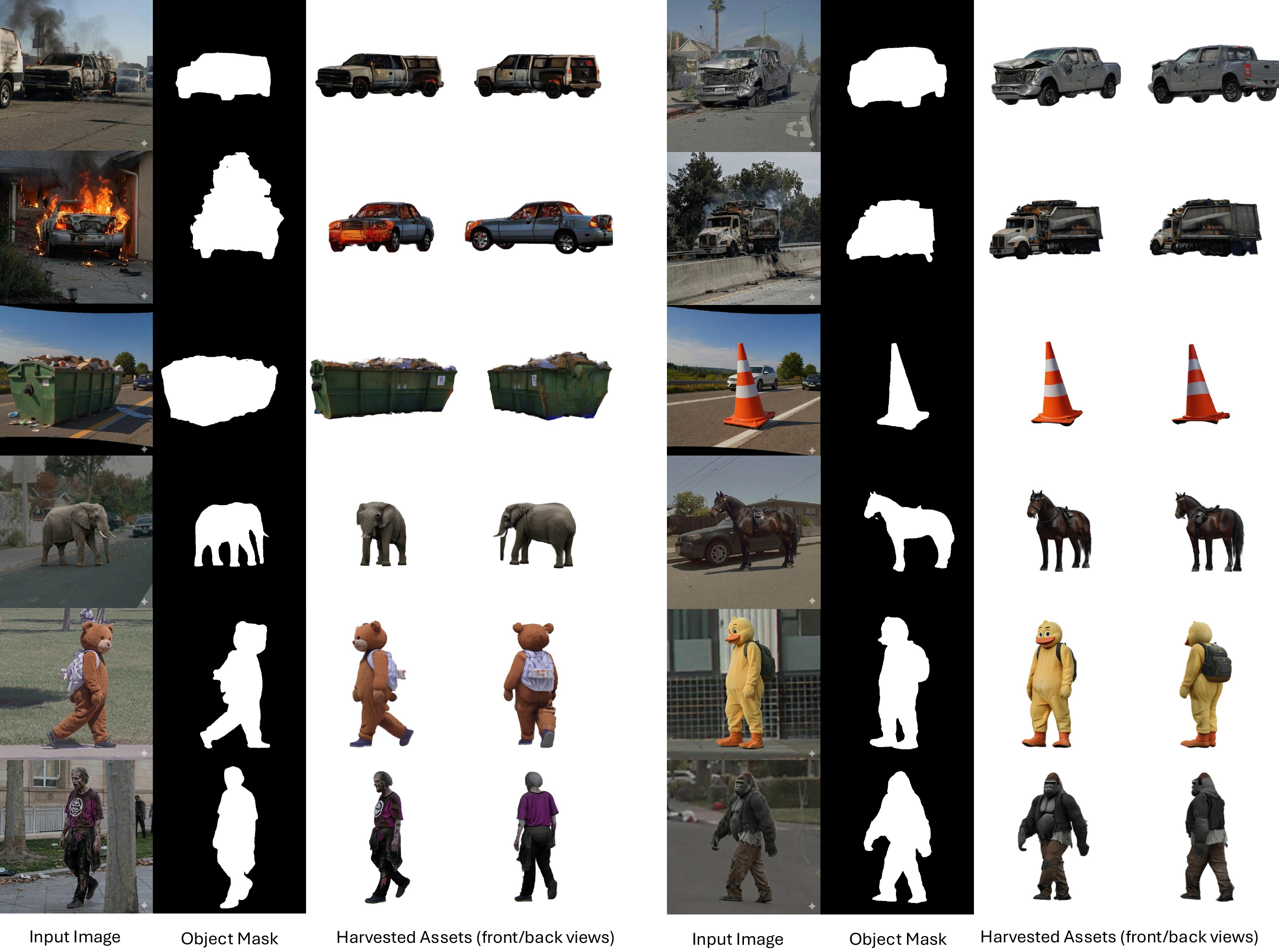}
\caption{OOD image editing and generalization with \PipeName. We edit AV inputs with Nano Banana to create out-of-distribution objects while preserving viewpoints, object poses, and scene background, then reconstruct plausible 3D assets from the edited observations.}
\label{fig:ood-generalization}
\end{figure*}

\begin{figure*}[t]
\centering
\includegraphics[width=0.9\textwidth]{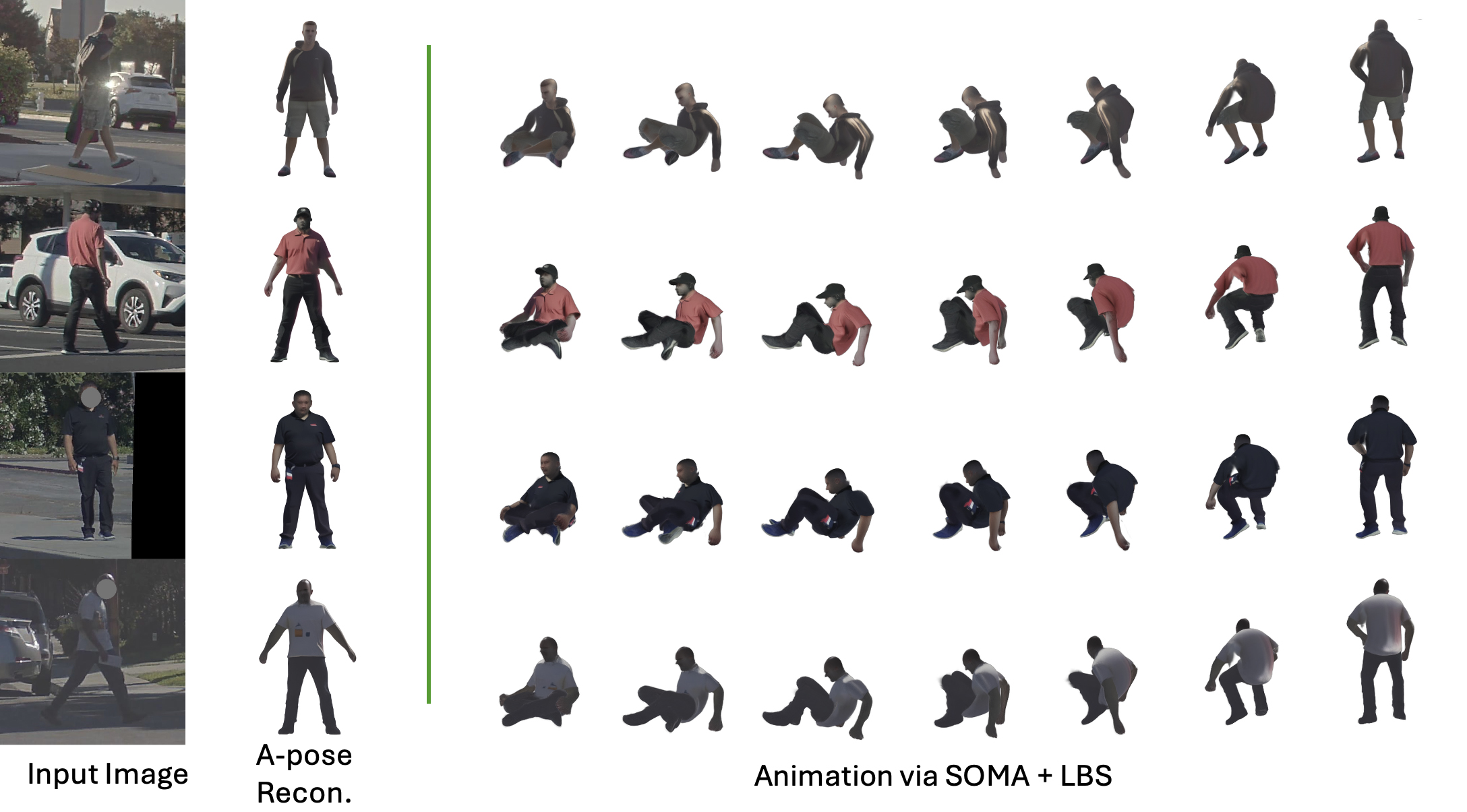}
\caption{Pedestrian animation results with \PipeName. We convert input observations into an A-pose asset, rig the generated asset with a simple LBS implemented with SOMA and GEM, and animate it with Kimodo.}
\label{fig:pedestrian-animation}
\end{figure*}

\begin{figure*}[t]
\centering
\includegraphics[width=0.9\textwidth]{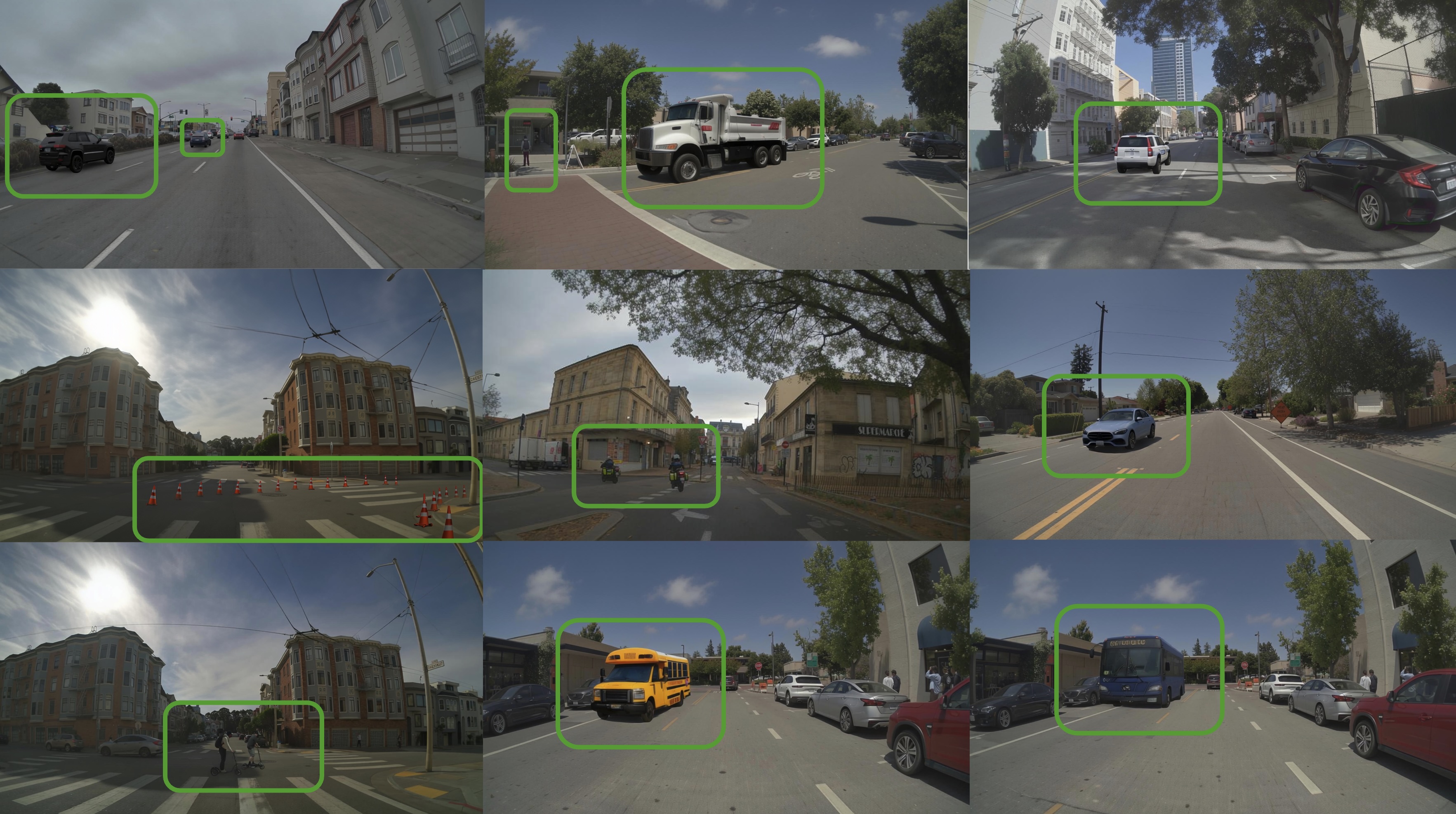}
\caption{Insertion and harmonization results with \PipeName. We reinsert generated assets into NuRec-reconstructed 3D scenes and apply DiffusionHarmonizer to reduce residual artifacts, improve shadows, and enhance local photometric consistency for better scene integration.}
\label{fig:insert-harmonization}
\end{figure*}

\subsection{Qualitative Results}

We provide representative examples for major object classes in Figure~\ref{fig:in-the-wild} and include challenging conditions commonly observed in autonomous-driving logs, such as occlusions, night scenes, human--object interactions,  blurry and low-resolution crops. Across these cases, \PipeName consistently produces plausible 3D assets with stable appearance.


\paragraph{Image Editing and OOD Generalization.}
Although \PipeName is developed for AV use cases and trained primarily on AV data, it still generalizes well to out-of-distribution (OOD) images. This generalization broadens \PipeName applications, enabling long-tail AV asset generation as well as flexible object editing and customization. In the examples shown in Figure~\ref{fig:ood-generalization}, we generate OOD images using Nano Banana~\cite{google2026gemini3flash} and prompt it to edit the object in AV images while preserving camera viewpoints, object poses, and background.

\paragraph{Pedestrian Animation}
For pedestrian animation, we first take the first input view and convert it to an A-pose image using Qwen-Image-Edit-2511~\citep{wu2025qwenimagetechnicalreport} with the prompt shown below.

\begin{center}
\fbox{\parbox{0.95\linewidth}{
Change the posture of the person in the image to A-pose: standing in a neutral A-pose, arms slightly lowered about 30 degrees from horizontal, legs shoulder-width apart. Do not change the position and camera viewpoint. Do not change the background, lighting, or the person's location in the frame. Maintain the aspect ratio and image quality.
}}
\end{center}

We then generate an A-pose pedestrian asset with our pipeline and rig it using a simple linear blend skinning (LBS) method implemented with NVIDIA SOMA~\citep{soma2026} and GEM~\citep{genmo2025}. Figure~\ref{fig:pedestrian-animation} shows representative animation results in which the human motion is generated by Kimodo~\citep{Kimodo2026}.

\paragraph{Insertion and Harmonization.}
We insert generated assets into 3D scenes reconstructed by NuRec. We further apply DiffusionHarmonizer~\cite{zhang2026diffusionharmonizer} to reduce residual artifacts and improve visual integration, including more coherent shadows and stronger local photometric consistency. Example results are shown in Figure~\ref{fig:insert-harmonization}.


\vspace{-2mm}
\section{Related Work}
\vspace{-2mm}

\paragraph{Image-to-3D models}

Recent image-conditioned 3D generation models aim to recover complete geometry and appearance from one or a few input views~\citep{sam3dteam2025sam3d3dfyimages,yang2024hunyuan3d,xiang2024trellis}.
SAM3D~\citep{sam3dteam2025sam3d3dfyimages} generalizes to in-the-wild images by training on large-scale
curated 3D data and reconstructed assets by itself, while Hunyuan3D~\citep{yang2024hunyuan3d,zhao2025hunyuan3d2,lai2025hunyuan3d25highfidelity3d} emphasizes high-quality textured asset creation with a design-oriented pipeline and typically cleaner inputs. TRELLIS and TRELLIS.2~\citep{xiang2024trellis,xiang2025trellis2} focus on structured latent representations for detailed geometry and appearance. 
Closely related to our method is a line of prior works~\cite{liu2023zero1to3,shi2023mvdream,wang2023imagedream,wu2024direct3d,gao2024cat3d} that leverages the general capabilities of text-to-image models to construct multi-view image generators as a core component of image-to-3D pipelines. Inspired by these works, \PipeName incorporates a multi-view image DiT and pairs it with a training strategy and data recipe tailored to extracting 3D assets from real driving logs.
Many of these methods are primarily evaluated on synthetic or curated benchmarks.
Autonomous driving logs, however, present sparse viewpoints, motion blur, occlusion, and calibration noise issues that motivate domain-specific asset extraction.

\paragraph{Driving-Data-Centric Asset Generation.}
GenAssets~\citep{yang2025genassets} and GINA-3D~\citep{shen2023gina3d} are closer to our setting in that they learn from real-world driving data. GenAssets uses a reconstruct-then-generate pipeline and learns optimization-based latent codes before training diffusion in the latent space, which limits the representation to those latents. GINA-3D similarly learns implicit neural assets from driving data with a tri-plane latent structure; as an implicit representation, we find it less suited for sharp surface detail compared to mesh/splat outputs.
Recent vehicle-focused methods further highlight complementary design points. DreamCar~\citep{du2024dreamcar} targets car reconstruction from very sparse forward-facing observations using car-specific generative priors, symmetry cues, and pose refinement. UrbanCAD~\citep{lu2024urbancad} emphasizes controllable and photorealistic CAD-based digital twins with retrieval-and-optimization for simulation-time editing and insertion. RGM~\citep{chen2024rgm} explores relightable 3DGS car asset generation from a single image by explicitly modeling materials and illumination. These methods underscore the value of domain priors, controllability, and relightability for driving simulation, while our focus is an end-to-end log-to-asset pipeline integrated with reconstruction-centric AV simulation.

\paragraph{Neural Reconstruction and AV Simulation.}
Neural reconstruction approaches such as NeRF~\citep{mildenhall2020nerf}, 3DGS~\citep{kerbl2023gaussian}, 3DGUT~\citep{wu20253dgut}, Dynamic Neural Scene Graphs (DNSG)~\citep{ost2021neural}, and OmniRe~\citep{chen2025omnire} show strong performance on novel-view synthesis and dynamic-scene reconstruction. However, these methods remain constrained by observed data and generally cannot complete unseen regions of scenes or objects. This creates a practical gap for closed-loop simulation systems such as NuRec~\citep{nurec_website}, CARLA~\citep{carla_nurec_docs}, and AlpaSim~\citep{alpasim_2025}, which require reusable and editable simulation-ready assets. Asset-Harvester is designed to bridge this gap.

\paragraph{Positioning of Our Work.}
Overall, we emphasize an end-to-end log-to-asset pipeline that targets simulation-ready outputs and practical deployment considerations, while designing the model to be tolerant of in-the-wild issues.
\vspace{-0.7em}
\section{Conclusion}
\vspace{-0.7em}

\label{sec:conclusion}
We presented \PipeName, a log-to-asset pipeline that converts sparse, noisy, and limited-angle observations from autonomous-driving logs into simulation-ready 3D assets. The system combines NCore-based object-centric data ingestion and curation, SparseViewDiT for sparse-view-conditioned multiview generation, and TokenGS-based feedforward lifting to recover complete geometry and appearance from challenging real-world inputs. Across the pipeline, hybrid real/synthetic/self-distilled training data and robustness-oriented preprocessing and augmentation help address occlusion, calibration noise, motion blur, and viewpoint bias in AV logs.
Experiments on the NuRec AV Object Benchmark show that \PipeName consistently outperforms strong image-to-3D baselines and remains effective in harder in-the-wild settings. We hope this work helps narrow the gap between neural reconstruction and practical closed-loop AV simulation, and provides a foundation for scalable asset extraction from real driving data.

\section{Acknowledgement}
\label{sec:acknowledgement}

We acknowledge the following colleagues (listed in alphabetical order by last name) for helpful discussions, feedback and support: Eric Cameracci, Bangbang Cao, Vincent Caux-Brisebois, Junsong Chen, Matt Cragun, Riccardo de Lutio, Yasaman Esfandiari, Janick Martinez Esturo, Song Han, Umar Iqbal, JF Lafleche, Jiefeng Li, Hai Loc Lu, Alex Perec, Sean Pieper, Ray Poudrier, Naveen Kumar Rai, Nick Schneider, Michal Tyszkiewicz, Qiao Wang, Enze Xie, Rurui Ye, Chang Yuan, Itai Zadok, Lei Zhang, Weihua Zhang, and Luis Zhu.

\appendix 

\clearpage
\appendix
\section{Per-class evaluation results}
\begin{table*}[h!]
\centering
\caption{ Quantitative comparison with image-to-3D baselines on Part A of the NuRec AV Object Benchmark. For each class, we compute mean over samples. PSNR in dB; ED-R is mean embedding distance.}
\label{tab:per-class-metrics}
\resizebox{\textwidth}{!}{%
\large
\begin{tabular}{l|cccc|cccc|cccc|cccc|cccc}
\toprule
Method & \multicolumn{4}{c|}{VRU pedestrians} & \multicolumn{4}{c|}{VRU riders} & \multicolumn{4}{c|}{commercial vehicles} & \multicolumn{4}{c|}{consumer vehicles} & \multicolumn{4}{c}{other objects} \\
\midrule
 & PSNR & LPIPS & SSIM & ED-R & PSNR & LPIPS & SSIM & ED-R & PSNR & LPIPS & SSIM & ED-R & PSNR & LPIPS & SSIM & ED-R & PSNR & LPIPS & SSIM & ED-R \\
\midrule
Trellis & 20.65 & 0.206 & 0.871 & 0.241 & 18.99 & 0.264 & 0.792 & 0.194 & 20.43 & 0.208 & 0.845 & 0.112 & 20.99 & 0.178 & 0.857 & 0.148 & 19.50 & 0.287 & 0.816 & 0.156 \\
Hunyuan3D-2 & 20.63 & \underline{0.190} & 0.873 & 0.295 & 18.40 & 0.246 & 0.789 & 0.205 & 19.21 & 0.216 & 0.838 & 0.157 & 19.58 & 0.194 & 0.851 & 0.197 & 18.25 & 0.266 & 0.808 & 0.159 \\
Hunyuan3D-2.1 & 19.85 & 0.234 & 0.859 & 0.355 & 17.51 & 0.306 & 0.760 & 0.356 & 19.18 & 0.223 & 0.826 & 0.222 & 18.83 & 0.199 & 0.825 & 0.206 & 17.78 & 0.306 & 0.790 & 0.232 \\
SAM3D & 21.20 & 0.194 & \underline{0.876} & 0.212 & 18.93 & \underline{0.236} & 0.788 & \underline{0.153} & 19.38 & 0.196 & 0.840 & \textbf{0.090} & 20.99 & 0.161 & 0.859 & 0.121 & 19.22 & \underline{0.254} & 0.813 & \textbf{0.104} \\
\midrule
\textbf{AH} (1V, est.~cam pose) & \underline{21.59} & 0.192 & \underline{0.876} & \underline{0.181} & \underline{19.46} & 0.247 & \underline{0.795} & 0.160 & \underline{21.48} & \underline{0.184} & \underline{0.849} & 0.094 & \underline{22.45} & \underline{0.147} & \underline{0.866} & \underline{0.090} & \underline{20.26} & 0.263 & \underline{0.819} & 0.151 \\
\textbf{AH} (1V, parsed cam pose) & \textbf{21.70} & \textbf{0.187} & \textbf{0.877} & \textbf{0.176} & \textbf{19.77} & \textbf{0.225} & \textbf{0.800} & \textbf{0.136} & \textbf{21.81} & \textbf{0.171} & \textbf{0.852} & \underline{0.093} & \textbf{22.69} & \textbf{0.137} & \textbf{0.869} & \textbf{0.083} & \textbf{20.72} & \textbf{0.240} & \textbf{0.825} & \underline{0.145} \\
\bottomrule
\end{tabular}
}
\end{table*}

\begin{table*}[htbp]
\centering
\caption{GPT-5.2 pairwise preference rates on Part B of the NuRec AV Object Benchmark. Higher percentages indicate more preferred results. In this experiment, Asset Harvester (AH) estimates camera parameters from a single input view with C-Radio linear probing.}
\label{tab:gpt-study-partb-perclass-estimatedcamera}
\small
\resizebox{\linewidth}{!}{%
\begin{tabular}{lcccccccc}
\hline
\textbf{Category} & \multicolumn{2}{c|}{\textbf{AH vs Trellis}} & \multicolumn{2}{c|}{\textbf{AH vs HY2}} & \multicolumn{2}{c|}{\textbf{AH vs HY2.1}} & \multicolumn{2}{c}{\textbf{AH vs SAM3D}} \\
\hline
 & AH (\%) & Baseline (\%) & AH (\%) & Baseline (\%) & AH (\%) & Baseline (\%) & AH (\%) & Baseline (\%) \\
\hline
consumer\_vehicles & \textbf{75.1} & 24.9 & \textbf{83.7} & 16.3 & \textbf{78.3} & 21.7 & \textbf{58.5} & 41.5 \\
commercial\_vehicles & \textbf{70.9} & 29.1 & \textbf{80.9} & 19.1 & \textbf{71.5} & 28.5 & \textbf{52.3} & 47.7 \\
VRU\_pedestrians & \textbf{79.5} & 20.5 & \textbf{85.8} & 14.2 & \textbf{82.3} & 17.7 & \textbf{65.2} & 34.8 \\
VRU\_riders & \textbf{73.3} & 26.7 & \textbf{80.0} & 20.0 & \textbf{80.0} & 20.0 & \textbf{64.4} & 35.6 \\
other\_objects & \textbf{65.8} & 34.2 & \textbf{63.4} & 36.6 & \textbf{72.0} & 28.0 & 46.1 & \textbf{53.9} \\
\hline
\textbf{average} & \textbf{72.9} & 27.1 & \textbf{78.8} & 21.2 & \textbf{76.8} & 23.2 & \textbf{57.3} & 42.7 \\
\hline
\end{tabular}
}
\end{table*}

\begin{table*}[htbp]
\centering
\caption{GPT-5.2 pairwise preference rates on Part B of the NuRec AV Object Benchmark. Higher percentages indicate more preferred results. In this experiment, Asset Harvester (AH) parses object camera from NCore scene for the single view input.}
\label{tab:gpt-study-partb-perclass-parsedcamera}
\small
\resizebox{\linewidth}{!}{%
\begin{tabular}{lcccccccc}
\hline
\textbf{Category} & \multicolumn{2}{c|}{\textbf{AH vs Trellis}} & \multicolumn{2}{c|}{\textbf{AH vs HY2}} & \multicolumn{2}{c|}{\textbf{AH vs HY2.1}} & \multicolumn{2}{c}{\textbf{AH vs SAM3D}} \\
\hline
 & AH (\%) & Baseline (\%) & AH (\%) & Baseline (\%) & AH (\%) & Baseline (\%) & AH (\%) & Baseline (\%) \\
\hline
consumer\_vehicles & \textbf{78.1} & 21.9 & \textbf{85.8} & 14.2 & \textbf{80.1} & 19.9 & \textbf{64.3} & 35.7 \\
commercial\_vehicles & \textbf{70.8} & 29.2 & \textbf{81.6} & 18.4 & \textbf{71.3} & 28.7 & \textbf{55.4} & 44.6 \\
VRU\_pedestrians & \textbf{82.7} & 17.3 & \textbf{87.7} & 12.3 & \textbf{83.7} & 16.3 & \textbf{70.5} & 29.5 \\
VRU\_riders & \textbf{76.7} & 23.3 & \textbf{73.3} & 26.7 & \textbf{75.6} & 24.4 & \textbf{64.4} & 35.6 \\
other\_objects & \textbf{61.3} & 38.7 & \textbf{63.4} & 36.6 & \textbf{67.5} & 32.5 & 44.4 & \textbf{55.6} \\
\hline
\textbf{average} & \textbf{73.9} & 26.1 & \textbf{78.4} & 21.6 & \textbf{75.6} & 24.4 & \textbf{59.8} & 40.2 \\
\hline
\end{tabular}
}
\end{table*}

\clearpage
{
\small

\bibliographystyle{unsrtnat}
\bibliography{neurips_2025}

}
\end{document}